\documentclass[sigconf]{acmart}
\PassOptionsToPackage{square,numbers}{natbib}
\usepackage{subcaption}

\AtBeginDocument{%
  \providecommand\BibTeX{{%
    \normalfont B\kern-0.5em{\scshape i\kern-0.25em b}\kern-0.8em\TeX}}}

\copyrightyear{2023} 
\acmYear{2023} 
\setcopyright{rightsretained} 
\acmConference[GECCO '23]{Genetic and Evolutionary Computation Conference}{July 15--19, 2023}{Lisbon, Portugal}
\acmBooktitle{Genetic and Evolutionary Computation Conference (GECCO '23), July 15--19, 2023, Lisbon, Portugal}\acmDOI{10.1145/3583131.3590355}
\acmISBN{979-8-4007-0119-1/23/07}

\begin{document}

\title{Optimization of a Hydrodynamic Computational Reservoir through Evolution}


\author{Alessandro Pierro}
\email{alessandropierro@proton.me}
\orcid{orcid.org/0000-0002-5682-627X}
\affiliation{
  \institution{Oslo Metropolitan University}
  \city{Oslo}
  \country{Norway} \\
  \institution{University of Trieste}
  \city{Trieste}
  \country{Italy}
}

\author{Kristine Heiney}
\email{kristine.heiney@oslomet.no}
\orcid{orcid.org/0000-0003-3332-4493}
\affiliation{%
  \institution{Oslo Metropolitan University}
  \city{Oslo}
  \country{Norway} \\
  \institution{Norwegian University of \\ Science and Technology}
  \city{Trondheim}
  \country{Norway}
}

\author{Shamit Shrivastava}
\email{shamit@apoha.ai}
\orcid{orcid.org/0000-0003-0916-7336}
\affiliation{%
  \institution{Apoha Limited}
  \city{London}
  \country{United Kingdom}
}

\author{Giulia Marcucci}
\email{Giulia.Marcucci@glasgow.ac.uk}
\orcid{orcid.org/0000-0001-7166-3643}
\affiliation{%
  \institution{Apoha Limited}
  \city{London}
  \country{United Kingdom} \\
  \institution{University of Glasgow}
  \city{Glasgow}
  \country{United Kingdom}
}

\author{Stefano Nichele}
\email{stefano.nichele@hiof.no}
\orcid{orcid.org/0000-0003-4696-9872}
\affiliation{%
  \institution{\O{}stfold University College}
  \city{Halden}
  \country{Norway} \\
  \institution{Oslo Metropolitan University}
  \city{Oslo}
  \country{Norway}
}


\begin{abstract}
As demand for computational resources reaches unprecedented levels, research is expanding into the use of complex material substrates for computing. In this study, we interface with a model of a hydrodynamic system, under development by a startup, as a computational reservoir and optimize its properties using an evolution in materio approach. Input data are encoded as waves applied to our shallow water reservoir, and the readout wave height is obtained at a fixed detection point. We optimized the readout times and how inputs are mapped to the wave amplitude or frequency using an evolutionary search algorithm, with the objective of maximizing the system's ability to linearly separate observations in the training data by maximizing the readout matrix determinant. Applying evolutionary methods to this reservoir system substantially improved separability on an XNOR task, in comparison to implementations with hand-selected parameters. We also applied our approach to a regression task and show that our approach improves out-of-sample accuracy. Results from this study will inform how we interface with the physical reservoir in future work, and we will use these methods to continue to optimize other aspects of the physical implementation of this system as a computational reservoir.\footnote{This work is carried out in part by co-authors S.~S.\ and G.~M.\ from Apoha, Ltd., a company developing hydrodynamic computational devices.}
\end{abstract}

\ccsdesc[500]{Computing methodologies~Evolvable hardware}

\keywords{complex systems, evolvable hardware, pattern recognition and classification}

\begin{teaserfigure}
  \includegraphics[width=\textwidth]{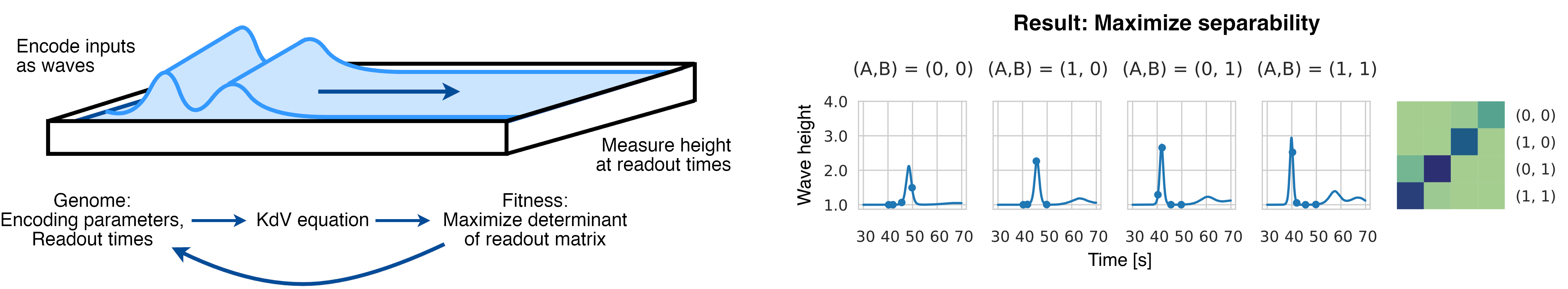}
  \caption{Computing with a hydrodynamic reservoir. We modeled a shallow water surface using the Korteweg--de Vries equation and used evolutionary computation to search for system parameters that maximize separability when this system is used as a computational reservoir.}
  \label{fig:teaser}
\end{teaserfigure}


\maketitle

\section{Introduction}
Complex dynamical systems, such as spiking neural networks (SNNs), cellular automata (CAs), and physical systems, have recently gained significant attention as potential alternatives to classical von Neumann architectures for computational tasks  \cite{Tanaka2019PhysResComp, adamatzky2016advances, konkoli2018reservoir}. In addition to being a topic of interest in their own right, the ability to efficiently harness these computational paradigms could lead to significant improvements in terms of time and energy consumption, as well as the development of new algorithmic solutions. 

The seminal yet provocative work ``\textit{Pattern recognition in a bucket}'' by \citeauthor{Fernando2003bucket} \cite{Fernando2003bucket} showed that it is indeed possible to use a bucket of water as \textit{liquid brain} substrate for computation. This was achieved by encoding inputs as nonlinear waves on the water surface. A camera was then used to capture the resulting waves after input interactions, and a simple linear layer was trained on top of the image data to solve the tasks at hand. 
While the work by \citeauthor{Fernando2003bucket} serves mainly as metaphor for the unconventional computing paradigm of liquid state machines (LSMs), the actual fabrication of computational devices based on the paradigm of liquid computation is of technological interest \cite{marcucci2023reservoir}.\footnote{The referenced work is carried out by Apoha Ltd., a company developing hydrodynamic computational devices. The work herein is carried out in part by co-authors S.~S.\ and G.~M.\ from Apoha Ltd.}

In this work, we focus on a common challenge in the design of these systems: the optimization of hyper-parameters. Specifically, we focus on the implementation of a physics-based reservoir model of a shallow water surface (Fig.~\ref{fig:teaser}); development of our modeled system as a physical reservoir is underway, and the simulation results obtained here will help advance this development. We benchmark this system on a nonlinear binary classification task and a regression task. We investigate the evolution of hyper-parameters using multi-dimensional archive of phenotypic elites (MAP-Elites) \cite{mouret2015illuminating} to identify a set of high-performing solutions. Our results show that it is beneficial to optimize the computation of liquid-based devices with evolutionary optimization. In particular, our solutions show better separability in a classification task than our previous implementation \cite{marcucci2023reservoir}, and when newly applied to a regression task, evolution produced solutions that improved out-of-sample accuracy. Our work paves the way to technological improvements enabling the future realization of liquid computation devices.

The remainder of this paper is organized as follows: in Section \ref{related-work}, we provide a brief review of related work. Section \ref{experimental} describes our experimental approach, including details of our specific case study and the computational paradigm we use. In Section \ref{results}, we present the results obtained on the XNOR and regression tasks. Finally, Section \ref{conclusion} offers some concluding remarks and suggestions for future work.

\section{Background and Related Work}
\label{related-work}

The aim of this study was to optimize the parameters of a physics-based reservoir model of a shallow water surface to yield optimal performance in the separability of the system readout.
This section will provide background on the main concepts behind this aim---reservoir computing and evolution in materio---and present related work done in this area.
Background on different optimization approaches is also presented to provide justification for the optimization strategies used here.

\subsection{Reservoir computing in simulo and in materio}

Reservoir computing is a framework designed to take advantage of the complex and nonlinear properties of a dynamical system, or \textit{reservoir}, for computation without the need for explicit training of the reservoir \cite{Schrauwen2007ReservoirComputing}.
In this framework, the reservoir maps inputs into a high-dimensional space, from which the target variables can be reconstructed with a trainable linear layer.

The benefit of this approach is that it allows for the use of systems with recurrent dynamics, which are well-suited for solving tasks with a temporal component, while sidestepping the need for the complicated procedures required to train such systems.
Reservoirs have been successfully applied to a number of tasks \cite{Schrauwen2007ReservoirComputing}, including controlling a simulated robot arm \cite{JoshiMaass2004robotarm}, speech recognition \cite{Skowronski2006speech, Verstraeten2005speech}, and chaotic time series prediction \cite{Jaeger2004MackeyGlass}.

More recently, reservoir computing has been expanded into the physical realm, with physical materials \cite{Tanaka2019PhysResComp} taking the place of the artificial and spiking neural networks (ANNs and SNNs) of the original echo state network \cite{Jaeger2001ESN} and liquid state machine \cite{maass2002real} frameworks.
Because reservoir computing is designed to take advantage of the dynamics inherent to the system, it is well-suited for exploiting the physical properties of materials.
With the right selection of material hardware for reservoir computing, rapid information processing can be achieved with little training overhead.

Physical systems that have been used as reservoirs are quite varied, ranging from biological \cite{Aaser2017,Sumi2022bnnreservoir} to electronic \cite{Jensen2017circuit} to spintronic \cite{Nakane2018}, to name a few.
Liquid systems have also been used for computation, both to build a fluidic analogue for electronic integrated circuits \cite{Thornsen2002,Draper2018,Prakash2007,Bartlett2022} and as computational reservoirs \cite{Fernando2003bucket}.
Of particular relevance to the present work is a study in which waves mechanically produced on the surface of a water bath were used as a reservoir \cite{Fernando2003bucket}.
Two tasks were performed in this study, XOR and speech recognition.
Inputs were encoded in the motion of eight motors driving small masses acting as point sources for the surface waves, and output images of the resulting wave interference patterns were fed into a single-layer perceptron to decode the output.
The authors demonstrated that the perceptron was better able to learn the speech recognition task when the inputs were first fed through the reservoir in this way, compared to a perceptron directly trained on the speech signals.

Our work builds on these demonstrations of the computing capacity of physical systems, approaching the following question:
Given a liquid system, governed by well-characterized physical laws (see Section \ref{setup-kdv}), can we tune the system parameters to optimize its performance as a computational reservoir?

\begin{figure}
    \centering
    \includegraphics[width=\linewidth]{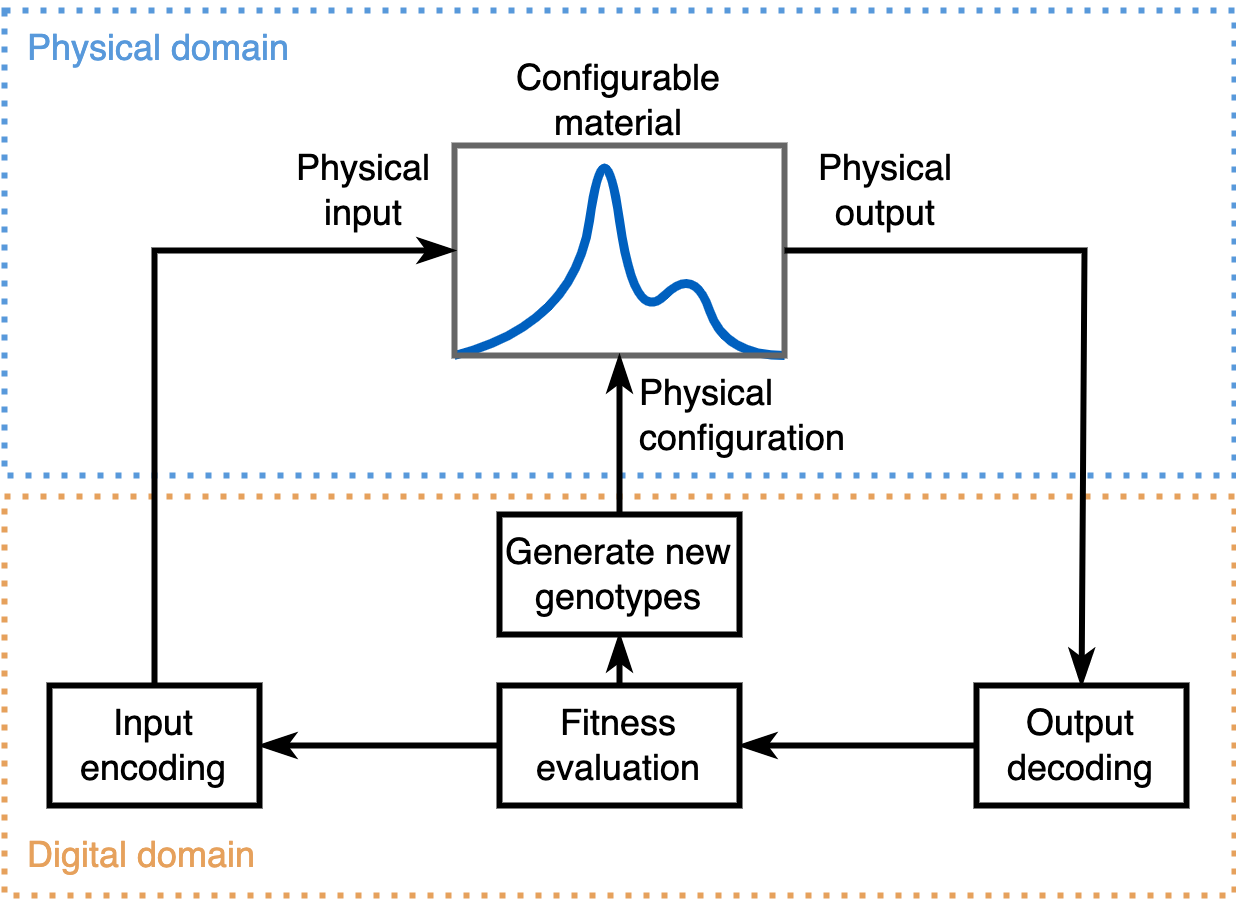}
    \caption{Architecture diagram for evolution in materio. Physical inputs and configurations of the target material are supplied by a computer. Physical outputs are read and decoded, and the decoded outputs are used to evaluate the fitness of the given inputs and configurations. New genotypes are obtained by a search algorithm, such as an evolutionary algorithm, and mapped to new inputs and configurations to test on the material. Adapted from \cite{Miller2014evoinmaterio}.}
    \label{fig:inmaterio-diagram}
\end{figure}

\subsection{Evolution in materio}

Although the use of physical materials for computation has many advantages, not all materials are well-suited for this application. Much in the same way that the connectivity of ESNs and LSMs must be tuned to produce dynamics rich enough to give inputs sufficiently separable representations in high-dimensional space \cite{Legenstein2007edgeofchaosLSM}, physical parameters of target materials must be programmed to yield dynamics suitable for computation. How then can we design material parameters to optimize computational task performance?

One approach to parameter optimization that has gained traction in recent years is \emph{evolution in materio} \cite{Miller2002evoinmaterio, Miller2014evoinmaterio}. In this approach, the evolutionary algorithm directly interfaces with the material to manipulate its intrinsic properties iteratively over generations to improve its performance on a target computational task (Fig.~\ref{fig:inmaterio-diagram}). With this approach, a large parameter space can be intelligently searched for high-performing solutions, and materials can be reconfigured as needed for different tasks.

In the current study, we developed an evolution in materio approach for a system consisting of a shallow water surface and applied it to a physics-based simulation of that system; in future work, we aim to use this approach to interface with the actual physical system and compare the real computational performance with our current predictions, thereby closing the \textit{reality gap}.

\subsection{Evolutionary and illumination algorithms}
\label{background-evo}

The selection of an effective hyper-parameter configuration is a crucial challenge in the design of reservoir computing systems. The high-dimensional nature of the search space, coupled with the nonlinear correlations among hyper-parameters, often make it difficult to use analytical or brute-force approaches for optimization. In order to address this challenge, simulation methods such as surrogate modeling \cite{yang2019surrogate, 10.1145/3205651.3205757} and the use of digital controllers for evolution in materio \cite{nichele2015investigation} can be useful in enabling efficient large-scale searches for effective hyper-parameter configurations.

Here, we focus on the standard version of multi-dimensional archive of phenotypic elites (MAP-Elites), formally introduced by Mouret and Clune \cite{mouret2015illuminating}.
In this algorithm, the feature space is discretized, and the highest-performing solution in each discrete region of the space is stored.
This illuminates the fitness landscape across the feature space, rather than aiming to find a single optimal solution.
With the final goal of translating our findings to the physical realization of our model system, this approach provides a number of benefits.
First, we can select the most physically implementable from a diverse set of high-performing solutions.
Second, we can target solutions in high-performing regions that span more of the search space to limit the sensitivity of the system's performance to implementation error.

\section{Experimental setup}
\label{experimental}

\begin{figure}
    \centering
    \includegraphics[width=\linewidth]{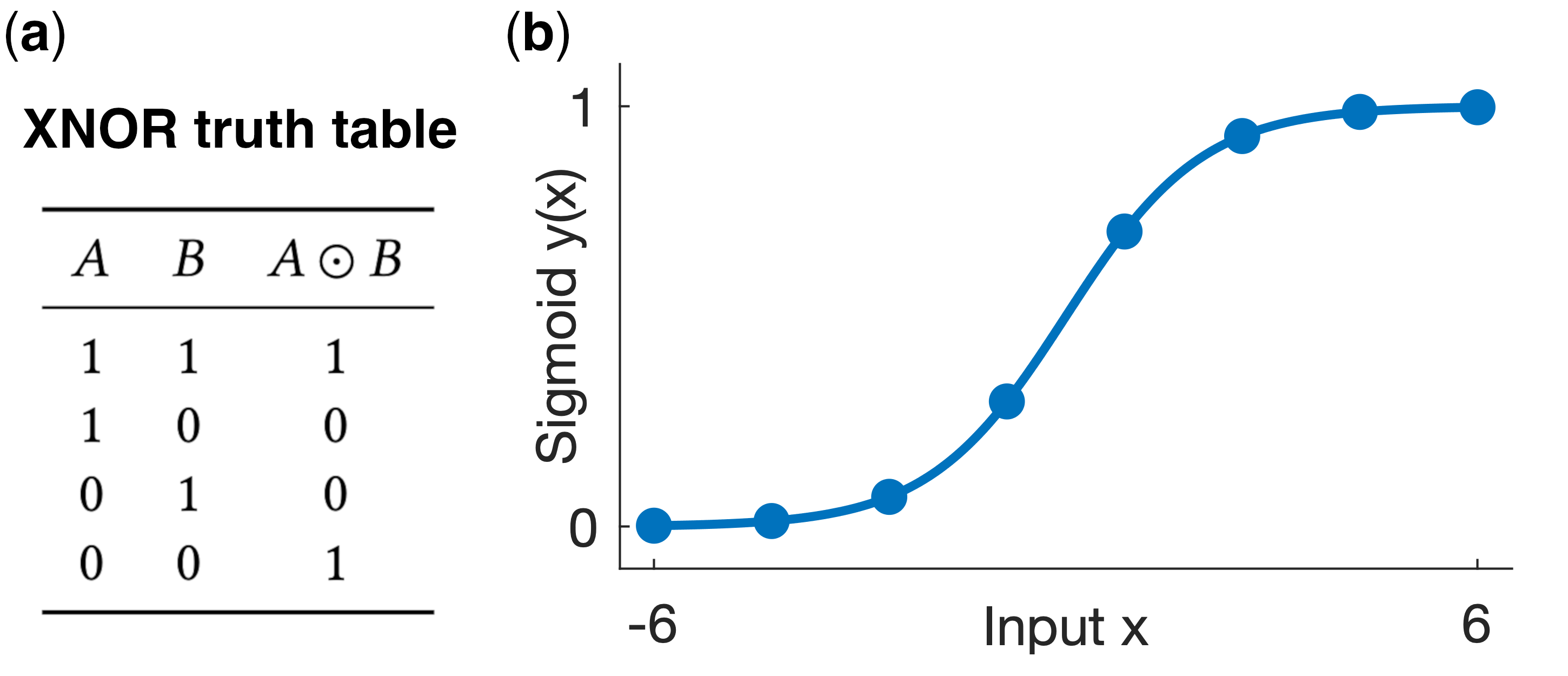}
    \caption{Two benchmark tasks used in this study. (a) Truth table for XNOR operator. (b) Sigmoid function used for the regression task. The dots represent the eight training data points.}
    \label{fig:sigmoid}
\end{figure}

The experimental setup and methodology of this study are presented in this section. Section \ref{setup-overview} gives an overview of the reservoir architecture and the target computational tasks used in this study. In Section \ref{setup-kdv}, we review the Korteweg--de Vries (KdV) equation, which forms the basis for our physics-based reservoir model. In Section \ref{setup-cnoidals}, we describe the encoding mechanism for input signals, both for discrete and continuous feature vectors. The readout step is outlined in Section \ref{setup-readout}, and in Section \ref{setup-fitness}, we explain how readings from the reservoir can be used to evaluate a given set of hyper-parameters. Finally, in Section \ref{setup-mapelites}, we present our evolutionary approach, based on the MAP-Elites algorithm.

\subsection{Pattern recognition with reservoirs}
\label{setup-overview}

As highlighted in Section \ref{related-work}, the nonlinear dynamics of reservoirs can be exploited as a powerful \emph{feature extraction} layer to solve pattern recognition problems, using a single trainable readout layer. We focused on supervised learning, with the aim of optimising the reservoir design to maximize the separability between individual observations in a data set. As will become clearer in the following sections, this objective is more general than the supervised setting, as achieving separability between different observations allows any target value assignment to be learnt at the additional cost of a single matrix multiplication. In order to focus our experimental setup, we adopted as benchmark tasks binary classification and univariate regression.


\subsubsection{Classification task}

Learning logic gates is a well-known formulation of binary classification from discrete variables. Considering the XNOR operator as an example (Fig.~\ref{fig:sigmoid}(a)), the goal is to compute $A \odot B$, given the values of $A$ and $B$. Although it is a simple problem, it would be impossible to solve with a single linear layer without exploiting the non-linearity of the reservoir, since the two binary classes are not linearly separable in the input space.

\subsubsection{Regression task}

The inherent analog nature of the reservoir can also be exploited to solve regression tasks with continuous variables. We focus on learning the sigmoid function,
\begin{equation}
    y(x) = \frac{1}{1+e^{-x}},
    \label{eqn:sigmoid}
\end{equation}
in the interval $[-6,6]$, based on $N=8$ equally spaced points, as shown in Fig.~\ref{fig:sigmoid}(b).
This task can be used to highlight how well the reservoir is able to generalize to out-of-sample data.

\begin{figure*}
    \centering
    \includegraphics[width=0.73\linewidth]{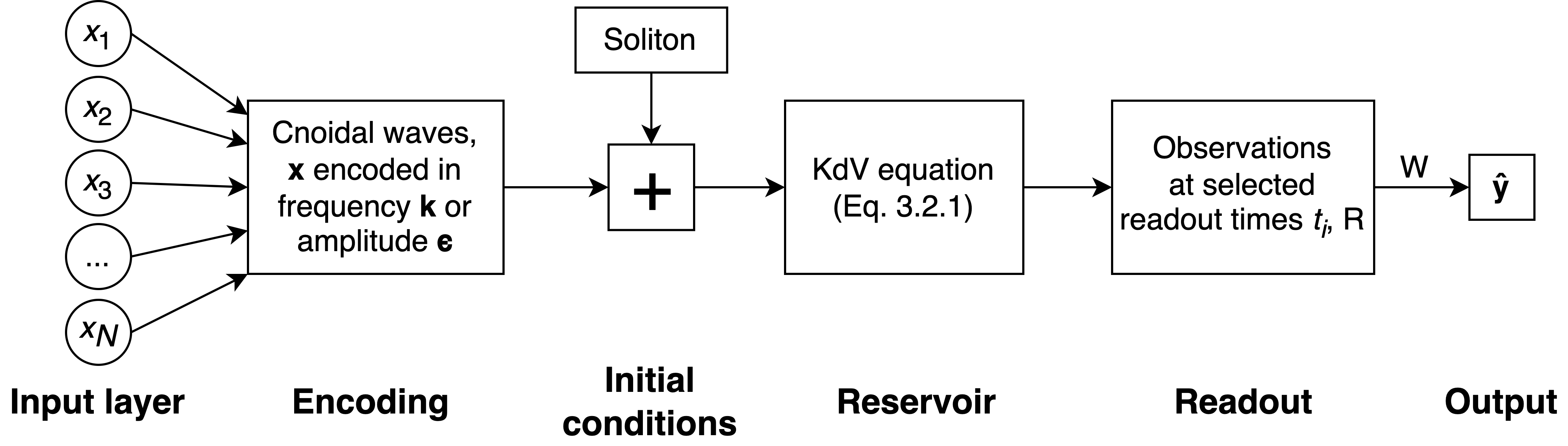}
    \caption{Architecture diagram. The input vector $\mathbf{x}$ is encoded in amplitudes $\mathbf{\epsilon}$ or frequencies $\mathbf{k}$ of cnoidal waves applied to the reservoir, along with a soliton. The matrix $R$ of observations at the readout times is converted to outputs $\hat{\mathbf{y}}$ via the weights $\mathrm{W}$.}
    \label{fig:diagram}
\end{figure*}

\subsubsection{Reservoir system architecture}

A sketch of the shallow water system we are simulating is shown on the left of Fig.~\ref{fig:teaser}, and the architecture used to interface with this system as a computational reservoir is shown in Fig.~\ref{fig:diagram}. We give an overview of the components here before explaining them in detail in subsequent sections.

The input vector $\mathbf{x}$ is encoded as amplitudes $\mathbf{\epsilon}$ or frequencies $\mathbf{k}$ of cnoidal waves for application to the shallow water reservoir. Details of the encoding approach are in Section \ref{setup-cnoidals}. Along with these input waves, a higher-amplitude soliton is also applied to provide baseline activity in the reservoir under zero-amplitude cnoidal input and enhance the complexity degree of the information processing stage, that is, provide nonlinearity in the wave mixing~\cite{marcucci2020theory}. Both the cnoidal and soliton waves extend uniformly across the width of the water surface and travel longitudinally along the pool (left to right in Fig.~\ref{fig:teaser}), propagating as described by the fluid dynamics model in Section \ref{setup-kdv}. This provides the nonlinear behavior required in a computational reservoir.

The displacement of the water at a distance $D$ from the inputs is then observed at a set of observation times $t_i$. Observations made for $N$ different inputs at these $M$ readout times gives an $N \times M$ readout matrix $\mathrm{R}$. A weight matrix $\mathrm{W}$ can then be used to convert $\mathrm{R}$ to output estimates $\hat{\mathbf{y}}$ of the target outputs $\mathbf{y}$. The readout process is described in more detail in Section \ref{setup-readout}.

The goal of this study is to maximize the reservoir's ability to separate inputs, by optimizing the encoding of inputs $\mathbf{x}$ as amplitudes $\mathbf{\epsilon}$ or frequencies $\mathbf{k}$ and the readout times $t_i$. The evolutionary approach used to accomplish this is presented in Section \ref{setup-mapelites}.

\subsection{Korteweg--de Vries equation}
\label{setup-kdv}

Fluids are some of the most complex systems studied in classical physics. Here, we focus on the Korteweg--de Vries (KdV) equation, a widely studied model that approximates the behavior of waves on shallow water surfaces \cite{drazin1989solitons}. The KdV equation is a nonlinear partial differential equation of the form:
\begin{equation}
\frac{\partial u}{\partial t} + u \frac{\partial u}{\partial \xi} + \lambda \frac{\partial^3 u}{\partial \xi^3} = 0,
\label{eqn:kdv}
\end{equation}
where $u$ denotes the height of the water, $t$ is time, $\xi$ is the spatial component, and the parameter $\lambda$ marks out the magnitude of the spatial dispersion.

One of the key features of the KdV equation is the presence of \emph{solitons}, which are wave solutions that retain their shape and velocity as they propagate. Solitons occur when the nonlinear term of the KdV equation (second term in Eq.~(\ref{eqn:kdv})) balances the KdV dispersion (third term in Eq.~(\ref{eqn:kdv})), and can be written as: 
\begin{equation}
    U_s(\xi, t) = U_0 + U_s \mathrm{sech}^2 \big[k_s\big(\xi-v_st\big)\big],
    \label{eqn:soliton}
\end{equation}
where $U_0$ denotes the water height at rest, and $U_s$, $k_s$, and $v_s$ represent the maximum height, wave number, and velocity of the soliton, respectively. In particular, the selection of the three coefficients $(U_s, k_s, v_s)$ has only two degrees of freedom to satisfy Eq.~(\ref{eqn:kdv}). For instance, the velocity can be determined from the other coefficients:
\begin{displaymath}
    v_s = U_0 +\frac{2}{3}U_s-4|\lambda| k_s^2.
\end{displaymath}

Our physics-based reservoir model is based on the KdV equation, and in particular on the behavior of Eq.~(\ref{eqn:kdv}) in a regular rectangular domain in $(1+1)D$ approximation; that is, each wave is stationary along the spatial direction perpendicular to the direction of travel $\xi$ (arrow in Fig.~\ref{fig:teaser}). Without any external force, the state of the reservoir at rest would result in a constant water level over the entire domain. We excite the reservoir at rest with a soliton wave, as described in Eq.~(\ref{eqn:soliton}), characterized by the following parameters: 
\begin{displaymath}
    U_0 = 1 \;\;\; U_s = 1 \;\;\; k_s = \frac{1}{2} \;\;\; v_s = \frac{4}{3}.
\end{displaymath}
The aim of this choice is to enhance the complexity of the reservoir and enable nonlinear interaction with the cnoidal waves used to encode the inputs, as we shall describe in the next section.

\subsection{Encoding datapoints as cnoidal waves}
\label{setup-cnoidals}

Let $\mathcal{D}=\{(\mathbf{x}_i, y_i)|i\in\mathcal{I}\}$ be a finite dataset of data points. In order to feed it  into the KdV reservoir, the feature vectors $\mathbf{x}_i$ must be encoded in signals that are compatible with Eq.~(\ref{eqn:kdv}). To achieve this, we focus on the use of \emph{cnoidal waves}~\cite{drazin1989solitons}, a family of solutions to the KdV equation that can be expressed as:
\begin{equation}
    u_c(\xi,t) = \epsilon_c \cos^2\big[k_c\big(\xi-v_ct\big)\big].
    \label{eqn:cnoidal}
\end{equation}
As in the case of the soliton, the parameters $(\epsilon_c,k_c,v_c)$ only have two degrees of freedom, with the velocity $v_c$ determined from the others as:
\begin{equation}
    v_c = U_0 + \frac{2}{3} \epsilon_c - 4 | \lambda | k_c^2.
    \label{eqn:vel-cnoidal}
\end{equation}
For the sake of generality, since every coordinate is adimensional, i.e., expressed in arbitrary units~(a.u.), we hereafter refer to the wavenumber $k$ as the frequency. Indeed, in this context, $k$ can be seen as a spatial frequency, while in different domains (e.g., soliton propagation in optical fibers), it can be a proper temporal frequency. 

We encode each observation in a superposition of $N_w$ cnoidal waves, with coefficients depending on the value assumed by $\mathbf{x}_i$. In particular, we shall consider the following two approaches.
\begin{itemize}
\item \emph{Amplitude encoding}: For each of the $N_w$ waves, a different, fixed, frequency is selected. The amplitudes $\epsilon_j$ are determined as functions of $x_i$. The velocities are then computed based on Eq.~(\ref{eqn:vel-cnoidal}):
    \begin{displaymath}
        \mathbf{x}_i\;\;\to\;\;\tilde{u}_i(\xi,t) = \sum_{j=0}^{N_w-1} \epsilon_j(\mathbf{x}_i) \cos^2\big[k_j\big(\xi-v_j(\mathbf{x}_i)t\big)\big].
    \end{displaymath}
\item \emph{Frequency encoding}: For each of the $N_w$ waves, a different, fixed, amplitude is selected. The frequencies $k_j$ are determined as functions of $x_i$. The velocities are then computed based on Eq.~(\ref{eqn:vel-cnoidal}):
    \begin{displaymath}
        \mathbf{x}_i\;\;\to\;\;\tilde{u}_i(\xi,t) = \sum_{j=0}^{N_w-1} \epsilon_j \cos^2\big[k_j(\mathbf{x}_i)\big(\xi-v_j(\mathbf{x}_i)t\big)\big].
    \end{displaymath}
\end{itemize}

The actual mapping from an observation $\mathbf{x}_i$ to its cnoidal encoding depends on the nature of its entries, particularly whether they are discrete or continuous. A given discrete entry with a finite domain $x_{i,j} \in \{0, \ldots, M-1\}$ is encoded in a single wave with a different encoding parameter for each possible value in the discrete domain. For example, adopting the amplitude encoding, $x_{i,j}$ would be encoded in a wave having fixed (but under evolutionary control) frequency $k$ and amplitude determined by the discrete value as:
\begin{equation}
    \epsilon(x_{i,j}) = \epsilon_{x_{i,j}} \;\;\;x_i \in \{0, M-1\}.
\end{equation}

\begin{figure}
    \centering
    \includegraphics[width=\linewidth]{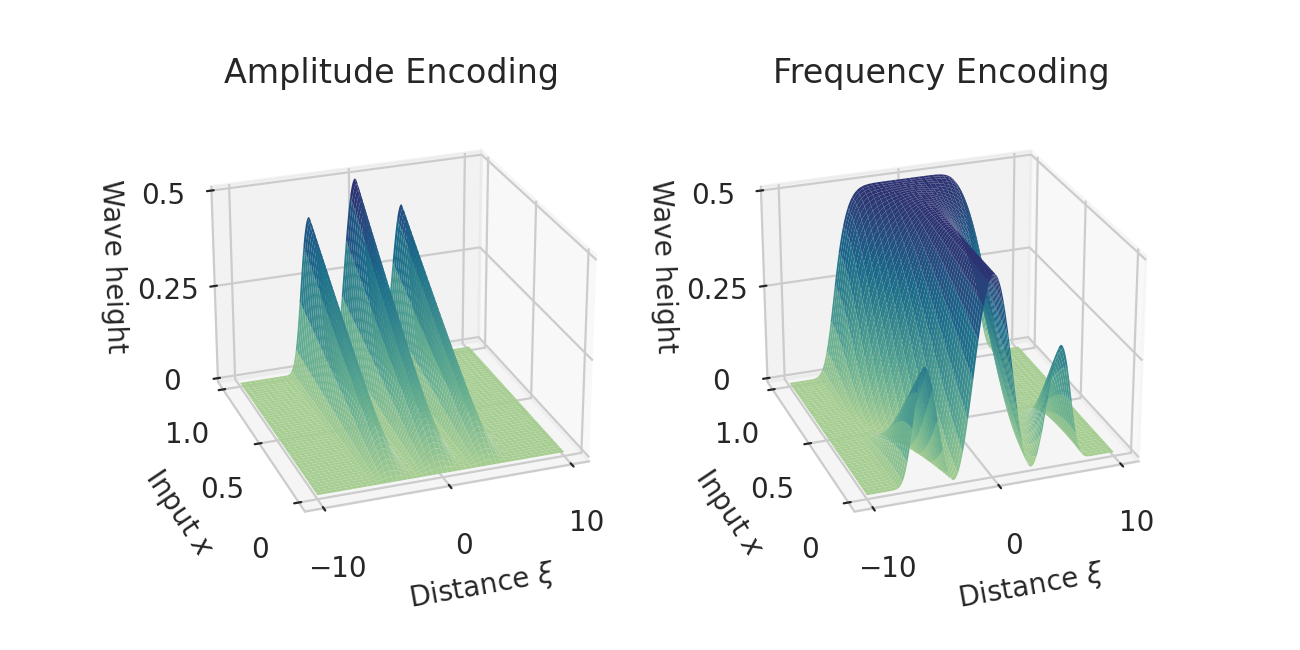}
    \caption{Wave profiles of cnoidal waves, generated with linear scaling, for amplitude encoding (left) and frequency encoding (right). A wave profile encoding a single input value corresponds to a slice of the surface with constant $x$.}
    \label{fig:encoding}
\end{figure}

In a similar fashion, continuous features are represented with fixed-point precision and a different wave is used to encode each digit. Based on the encoding approach, the amplitude or the frequency is determined to be proportional to the value of each digit, with the proportionality parameter under evolutionary control. For example, assuming $3$-digit precision and amplitude encoding, a continuous entry $x_i$ would be encoded in three waves with different frequencies $k_0, k_1, k_2$ and amplitudes proportional to the three digits $x_i^{(0)},x_i^{(1)},x_i^{(2)}$ as:
\begin{equation}
    \epsilon_j(x_i^{(j)}) = w_j \cdot x_i^{(j)}
\end{equation}
with all the weights $w_0,w_1,w_2$ appropriately bounded to ensure the resulting amplitude is in a physically feasible domain. Fig.~\ref{fig:encoding} shows how linearly scaling the amplitude (left) and the frequency (right) results in different wave profiles. Here, the wave corresponding to the encoding of a single digit is a slice of the surface along a line of constant \emph{Input} value $x$.

To ensure that the signals assume non-zero values only on a bounded domain, the resulting input signals are scaled with a super-Gaussian of eighth order, thus giving the following form: 
\begin{equation}
    u_j(x, t) = \tilde{u}_j(x,t)\exp\left[-\left(\frac{x}{l}\right)^8\right],
    \label{eqn:encoding}
\end{equation}
where we set the scale as $l=20$.

\subsection{Reading observations from the reservoir}
\label{setup-readout}

Given an encoding scheme and initial conditions augmented with the soliton, the reservoir can be simulated for an arbitrary amount of time. However, choosing the right timing for reading out observations plays a crucial role in the final performance of the model, as it can be easily seen that the encoding and timing choices are highly coupled. We choose to let this aspect under evolutionary control, adding to the genotype a list of instants of time as:
\begin{displaymath}
    0\leq t_0 \leq \ldots \leq t_r \leq T_{max}.
\end{displaymath}

We organize the set of readings from the reservoir in a \emph{readout matrix} $R$, where the $r_{i,j}$ is the $j$th readout value associated with the $i$th observation from the chosen dataset. We restrict our discussion to the case of a square readout matrix, choosing a number of readouts $r$ equal to the number of samples in our training dataset.

\subsection{Fitness function to evaluate separability}
\label{setup-fitness}

Let $\mathcal{H}$ be the set of hyper-parameters, i.e., the genotype of the evolutionary algorithm, composed of the parameters pertaining to the encoding step and the readout times:
\begin{equation}
    \mathcal{H}=(t_0, \ldots, t_k, \epsilon_0, \ldots, \epsilon_t, k_0, \ldots, k_p)\;\;.
\end{equation}
Our objective is to optimize the performance of the system on the presented supervised learning tasks. In particular, the output values can be reconstructed through a linear layer as:
\begin{equation}
    \hat{\mathbf{y}} = WR+\mathbf{b}\;\;,
\end{equation}
where $W$ is the matrix of weights converting the readout matrix $R$ to the readouts $\mathbf{y}$.
Setting $\mathbf{b}=\mathbf{0}$, we obtain:
\begin{equation}
    W = \mathrm{argmin}_W \|\hat{\mathbf{y}}-\mathbf{y}\|_2\;\;.
\end{equation}

If $R$ is a square, full-rank matrix, $W$ can be computed in closed form as $W=\mathbf{y}R^{-1}$. However, a full-rank readout matrix is difficult to obtain without tuning the hyper-parameters, especially as the number of training data points increases. Thus, we focus on maximising the determinant of the readout matrix $R$ as a measure of non-singularity measure, defining the following objective function:
\begin{equation}
    F(\mathcal{H}; \mathcal{D}) = | \det R |.
\end{equation}

It should be emphasized at this point that our goal is to maximize the system's ability to linearly separate each of the observations in the dataset, irrespective of the class label associated to it. This task is more general than simple classification. In fact, during the training and hyper-parameter tuning phases, the target variables are not even used. Given a high-fitness hyper-parameter configuration and the resulting weight matrix $W$, a prediction can be computed for any point by running its associated simulation, retrieving the readout form the reservoir, and multiplying the resulting vector for the weights $W$. Results in Section \ref{results-fit_validation} supports the effectiveness of this fitness at improving the accuracy of the model on the test set.

\begin{figure}
\centering
     \begin{subfigure}[b]{\linewidth}
     	\centering
     	\includegraphics[width=0.9\linewidth]{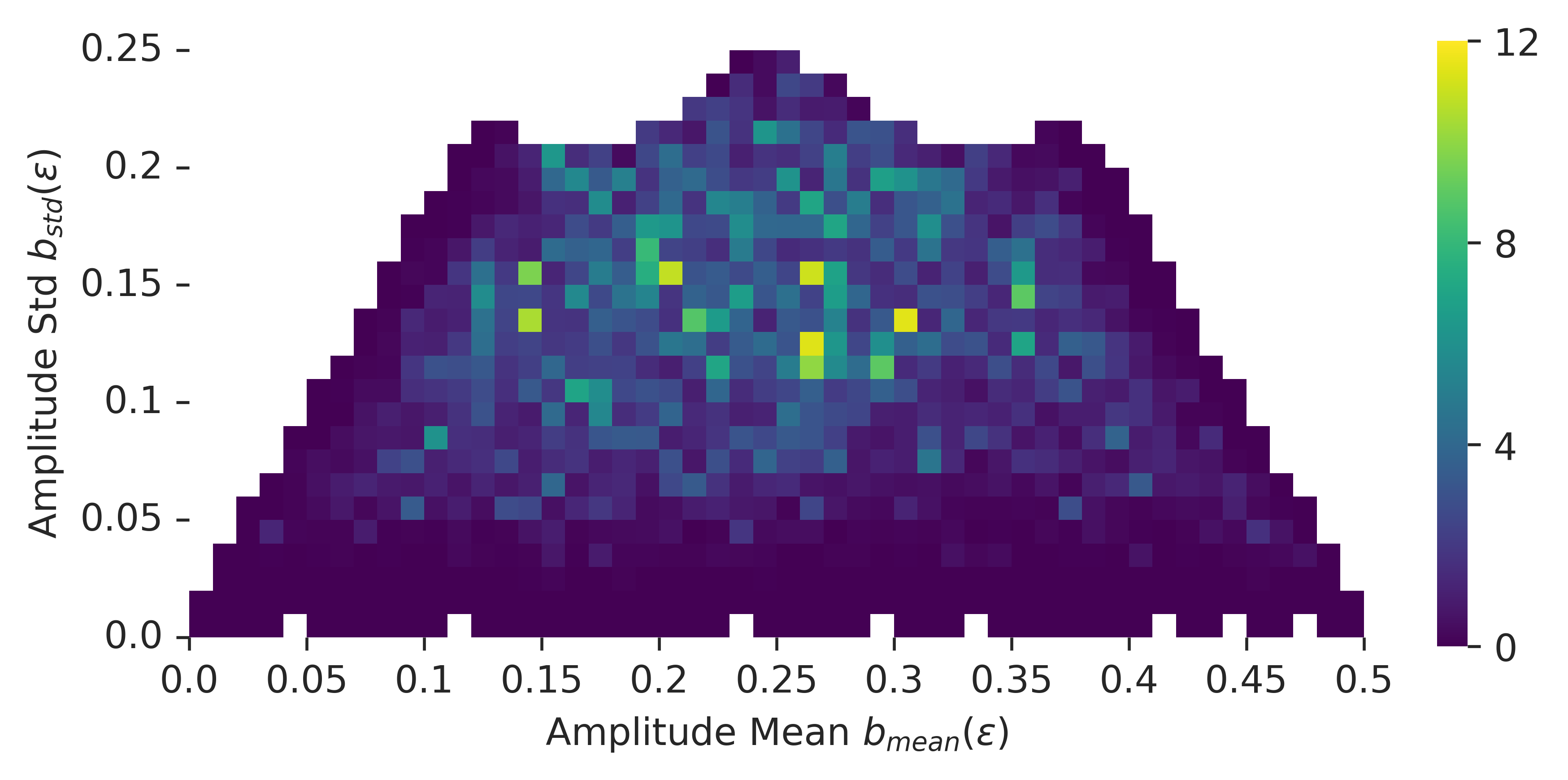}
     	\caption{Amplitude encoding}
     \end{subfigure}\\
     \begin{subfigure}[b]{\linewidth}
     	\centering
    	\includegraphics[width=0.9\linewidth]{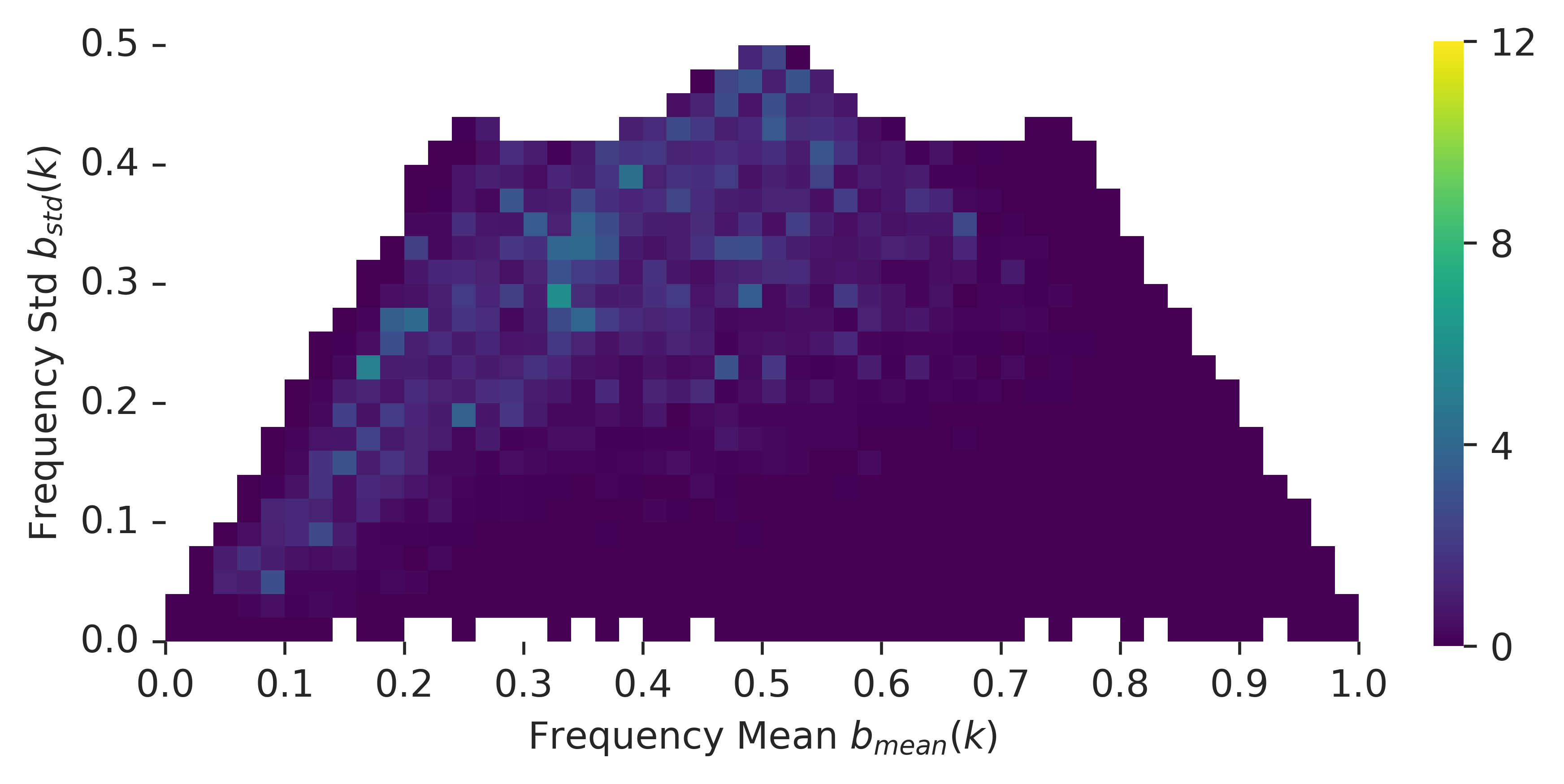}
     	\caption{Frequency encoding}
     \end{subfigure}\\
     \begin{subfigure}[b]{\linewidth}
     	\centering
    	\includegraphics[width=0.6\linewidth]{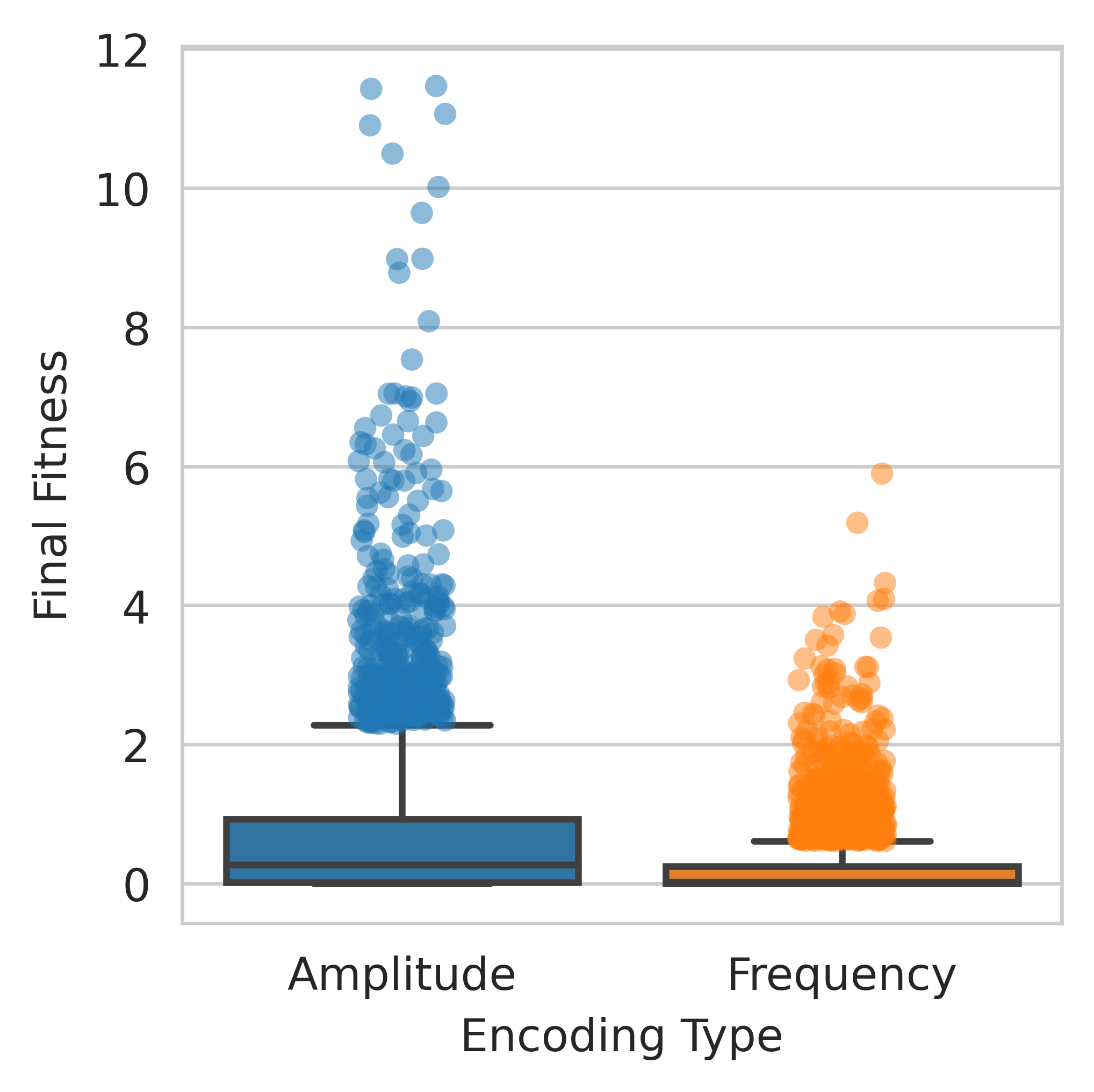}
     	\caption{Fitness distributions}
     \end{subfigure}
     \caption{XNOR task. Final MAP-Elites fitness archives for (a)~amplitude and (b) frequency encoding. The intensity of the heatmap represents the fitness. (c) Distribution of fitness values in the archives for amplitude and frequency encoding.}
     \label{fig:xnor-heatmap}
\end{figure}

\subsection{MAP-Elites algorithm}
\label{setup-mapelites}

A preliminary investigation using $(\mu, \lambda)$-Evolution Strategies pointed out that the symmetry in the encoding scheme and the non-convexity of the problem would have benefited from a more advanced evolutionary algorithm. This motivated us to base our approach on the MAP-Elites algorithm, as described in \cite{mouret2015illuminating}. The MAP-Elites explores a diverse set of solutions in a high-dimensional space by maintaining a map of the best solutions found so far across a grid of \emph{phenotypic descriptors}. To evolve a population of encodings operating in different regimes in terms of wave profiles, we chose to characterize our solutions using the mean and standard deviation of encoding wave parameters. For example, for the amplitude encoding, the phenotypic map would be based on the descriptors:
\begin{equation}
    \mathbf{b}(\epsilon) = \big(b_{mean}(\epsilon), b_{std}(\epsilon)\big)
\end{equation}
where $\epsilon$ is a vector containing all the cnoidal amplitudes observed during the fitness evaluation of a certain individual. 

We performed mutation with additive Gaussian noise $\mathcal{N}(\sigma=0.1)$ and scaled all variables linearly to the $[0,1]$ interval. 
All code to reproduce the experiments is available online.\footnote{\url{https://github.com/AlessandroPierro/HydrodynamicReservoirEvolution}}

\section{Results and Discussion}
\label{results}

In this section, we present and discuss the results of our experimental campaign. To begin, we evaluate the evolvability of the proposed fitness function and compare the effectiveness of the two encoding schemes on the benchmark task of learning the XNOR logic gate. Then, we apply our approach to the regression of a sigmoid function, showing how maximizing the separation capability of our system is beneficial for its accuracy on out-of-sample data points.

\begin{figure*}[h]
     \centering
     \includegraphics[width=0.7\linewidth]{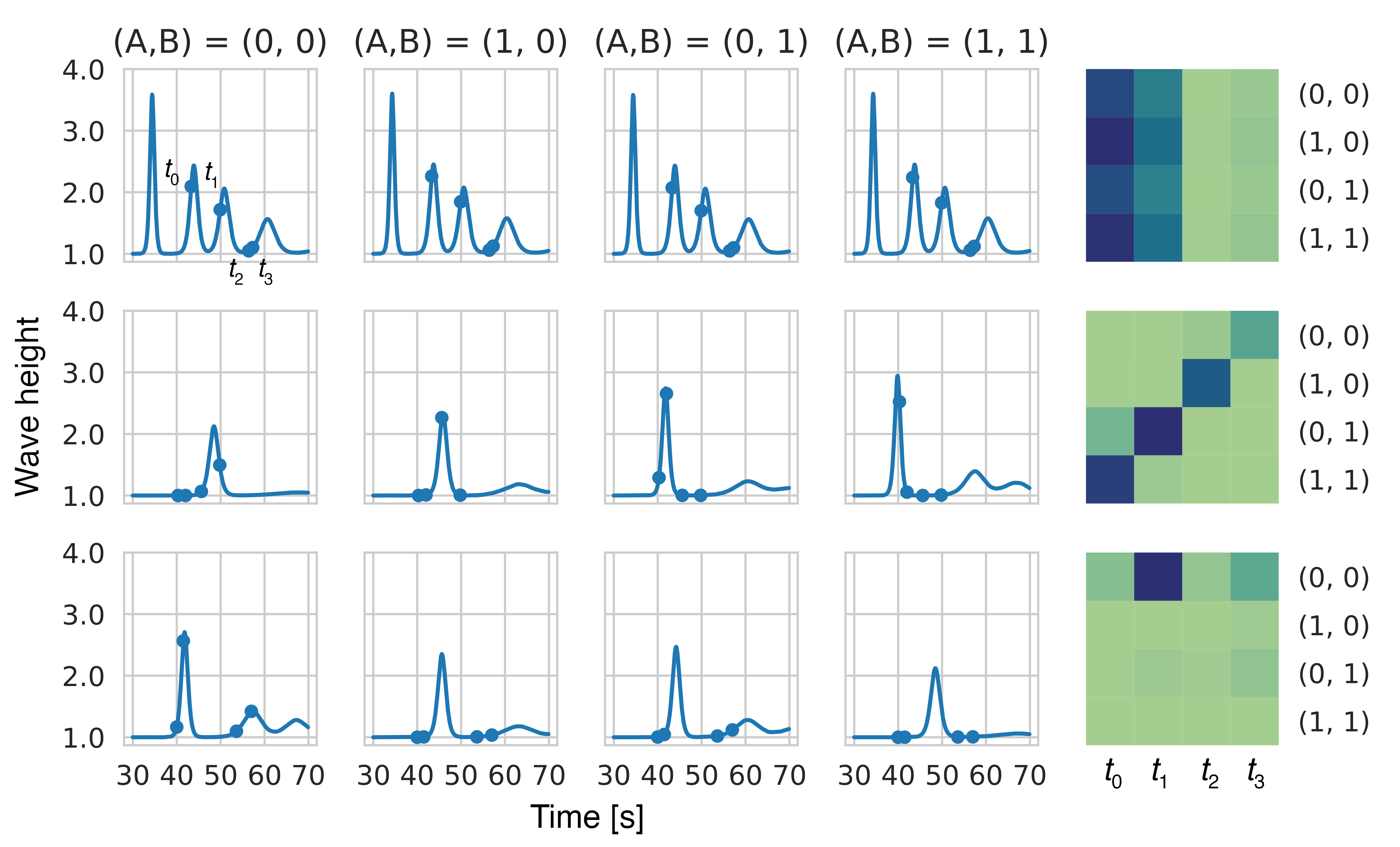}
     \caption{XNOR task. Wave profiles generated from three different encodings: a random encoding (top row), the highest-fitness evolved amplitude encoding (middle row), and the highest-fitness evolved frequency encoding (bottom row).}
     \label{fig:grid-waves}
\end{figure*} 

\subsection{Validation with logic gate learning}

Using the setup described in the previous section, we conducted two sets of experiments using amplitudes and frequencies for input encoding. We used the MAP-Elites algorithm and performed 10,000 fitness evaluations for each set of experiments. The results are summarized in Fig.~\ref{fig:xnor-heatmap}. All data are aggregated from five experimental runs with different random seeds.

\subsubsection{Amplitude works better than frequency}


A notable outcome of our experiments is that the amplitude encoding performs better than the frequency encoding, both in terms of the overall fitness landscape and in the maximum fitness achieved, as shown in Fig.~\ref{fig:xnor-heatmap}. 
Figure~\ref{fig:xnor-heatmap}(c) shows the distributions of the fitness values in the final archives for the two types of encoding.
As seen from these distributions, the amplitude encoding yielded significantly better fitness than the frequency encoding
(Mann–Whitney test, $p<10^{-167}$, $U=1.099\cdot10^7$, sample sizes $n_{ampl}=n_{freq}=4023$). 

This result is consistent with the known physics of cnoidal waves, which are more sensitive to changes in amplitude than frequency. Additionally, the results show that our approach yields solutions that greatly outperform our previous proof-of-concept, presented by Marcucci et al.~\cite{marcucci2023reservoir}.
In this previous study, the obtained solution had a readout matrix determinant of $|\det(R)| = 0.0115$, whereas our top performing solution had $|\det(R)| = 11.8$, a full three orders of magnitude greater.
This greater determinant yields more separable outputs, which was the goal of our evolutionary strategy.
This improved separability will be demonstrated by exploring specific solutions in the following section.

\subsubsection{Assessing separability in readouts}

To demonstrate the separability performance of the highest-fitness solutions obtained in with the two encoding methods, we show the outputs obtained with a random encoding and the top-performing solutions of each of the encoding methods in Fig.~\ref{fig:grid-waves}.
The line plots show the waveforms at the readout location in response to the four possible input combinations of A and B. The waveforms are plotted over time with the four readout times marked as dots.
The heatmaps are the readout matrices, with each row representing the readout wave heights for a given input.

The plot reveals that the random encoding (top row) leads to nearly identical profiles for the different data points, yielding low separability and low fitness. In contrast, the optimal solutions for each of the encoding methods show a diversity of output representations of the different inputs, particularly in the case of the amplitude encoding. Additionally, for the amplitude encoding, the coupled evolution of the wave parameters and the readout times resulted in a clear pattern in the readouts, with each data point being associated with a large wave peak at a different readout time.

\subsection{Application to regression task}
\label{results-fit_validation}

\begin{figure}[h]
     \centering
     \begin{subfigure}[b]{\linewidth}
     	\centering
     	\includegraphics[width=0.85\linewidth]{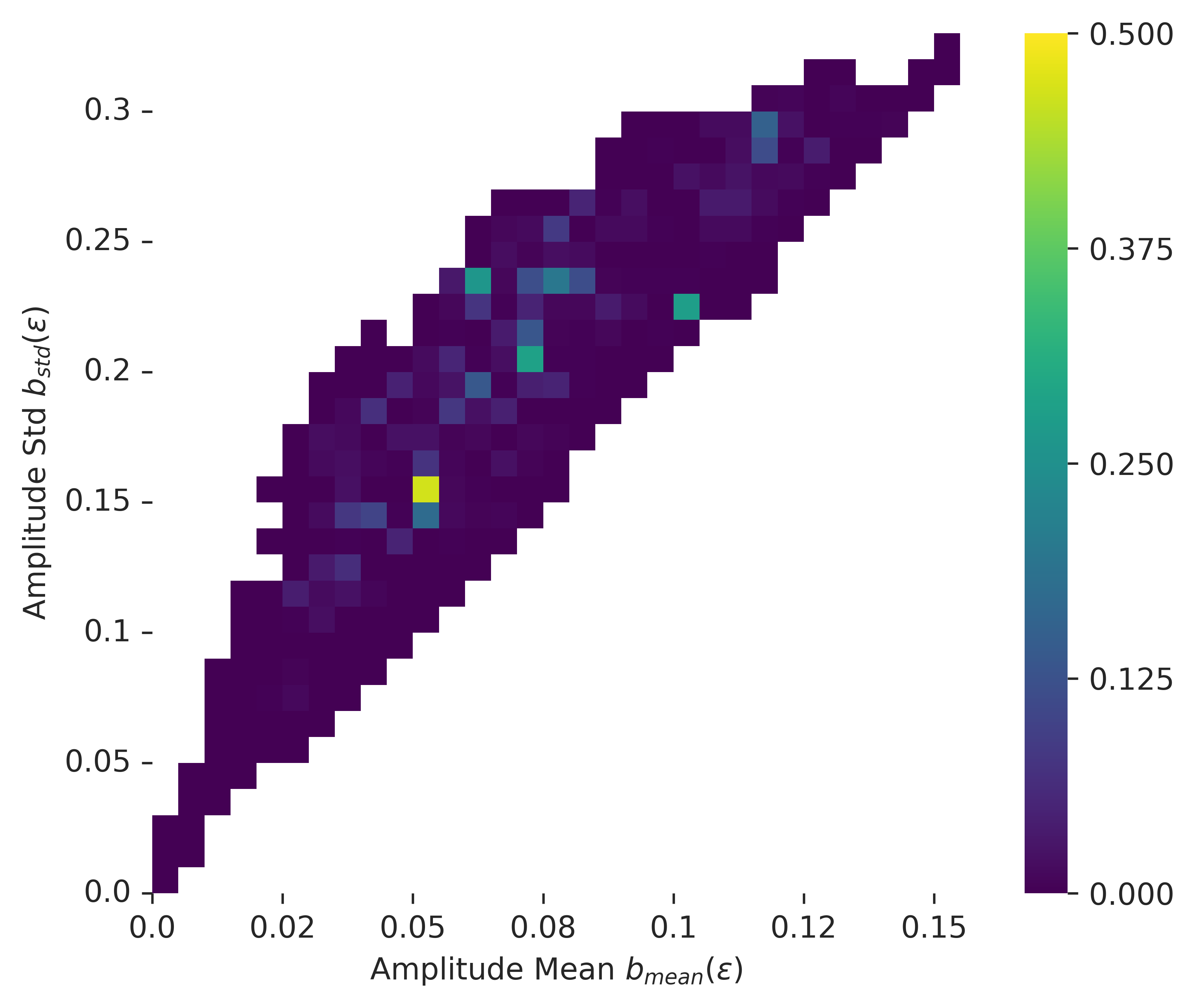}
     	\caption{Fitness of final archive}
     \end{subfigure}
     \begin{subfigure}[b]{\linewidth}
     	\centering
	\includegraphics[width=0.75\linewidth]{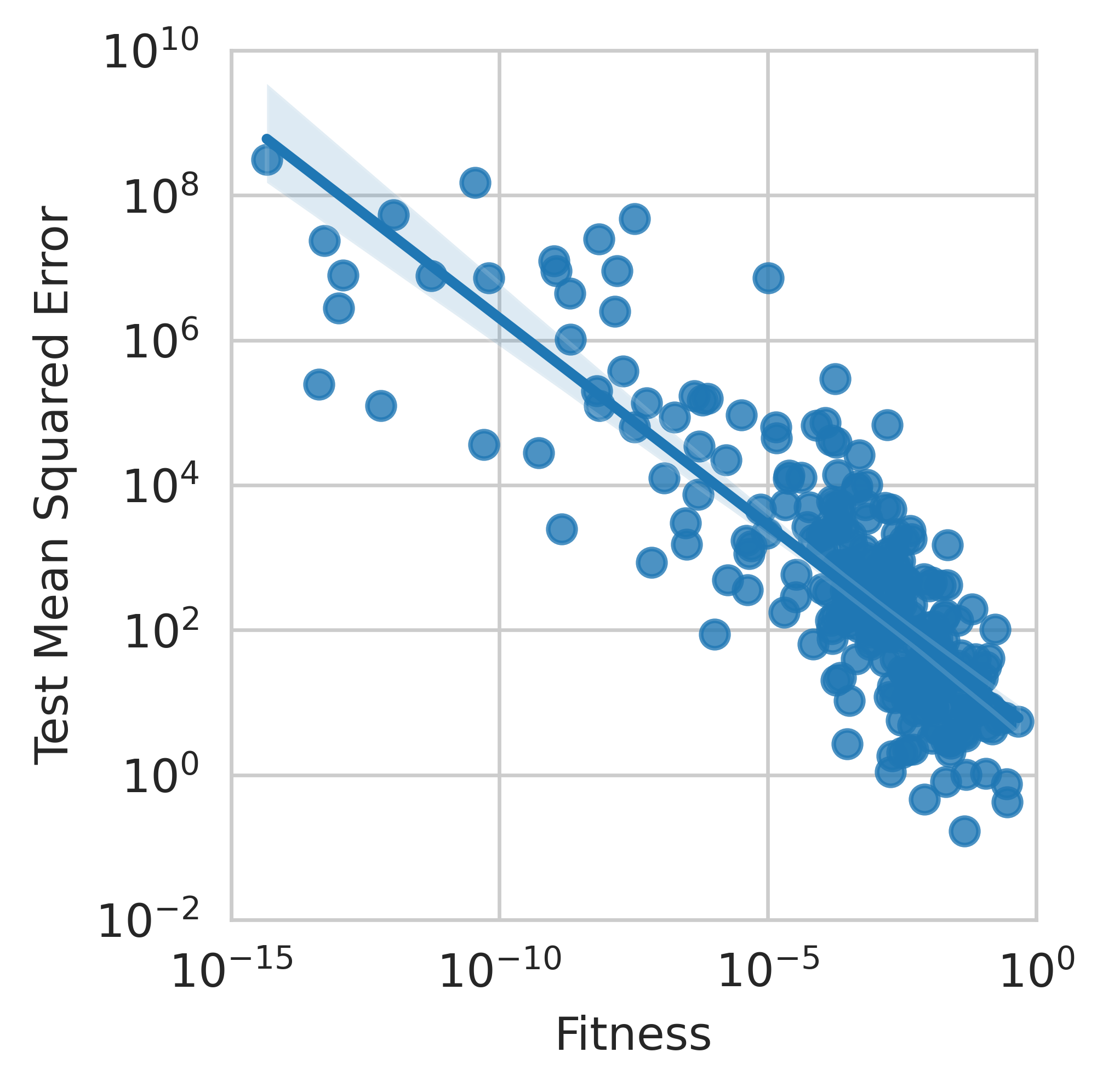}
    	\caption{Accuracy of test data}
     \end{subfigure}
     \caption{Regression task. (a) Final MAP-Elites fitness archive. The intensity of the heatmap represents the fitness. (b) Mean squared error of test data plotted against fitness for the final archive.}
     \label{fig:heatmap-regression}
\end{figure}

To test the geometrical generalization capabilities of our system and its ability to learn from subsets of training data, we conducted a set of experiments on a simple regression task. We focused on learning the sigmoid function in the interval $[-6, 6]$, based on a set of $N=8$ uniformly spaced points (Fig.~\ref{fig:sigmoid}(b)). We applied the amplitude encoding for continuous features, as described in Sec.~\ref{setup-cnoidals}, and performed 5,000 iterations of the MAP-Elites algorithm. The final archive is shown in Fig.~\ref{fig:heatmap-regression}(a).

We then evaluated how well our approach and fitness function serve to improve the accuracy of the evolved solutions on out-of-sample data.
For each hyper-parameter configuration in the final archive, we assessed the accuracy of the resulting solution in terms by calculating the mean squared error (MSE) for a set of $200$ test points, sampled uniformly in the same domain as the training set. The result is shown in Fig.~\ref{fig:heatmap-regression}(b), highlighting the correlation between fitness and accuracy.
The obtained linear regression model in log--log space is statistically significant ($R^2=0.686$, $F(1,269)=586.6$, $p<10^{-68}$).

This demonstrates first that our chosen fitness objective of maximizing the determinant of the readout matrix is able to drive the system toward higher-performing solutions.
Second, our results also demonstrate that our evolved solutions are able to generalize to test data within the input domain.
This is especially important considering our goal of maximizing separability of the inputs; our encoding method combined with the nonlinear dynamics inherent to the system are able to strike a balance between maximally separating the test inputs and still maintaining low distances between similar inputs in representational space.
This is a valuable finding in the context of larger data sets of continuous inputs: maximizing the separability given only a limited subset of training data can improve the test accuracy with few model evaluations.

However, one issue is that the MSE of our high-fitness solutions is still somewhat large. We hope to address this by using more complex models or nonlinear readouts, or by refining the encoding method.
Crucially, our fitness function emphasizes only the \textit{separation} property of the reservoir, whereby distinct inputs produce distinct outputs, and neglects to encourage the \textit{approximation} property, which leads to similar outputs in response to similar inputs \cite{maass2002real}.
In future work, we will consider fitness functions that can strike a balance between these two properties, with the high separability achieved here as a starting point.




\section{Conclusions and future work}
\label{conclusion}

Our work demonstrates the utility of an evolution in materio framework to optimize the performance of models of physical substrates for computation, and we will expand on our findings by inserting the physical system into this framework, in place of the model implemented here.
We applied MAP-Elites to obtain optimized encoding parameters and readout times that maximize the ability of a model shallow water reservoir to linearly separate training inputs, as measured by the determinant of the readout matrix.
Our experiments were based on a simulated model in place of the actual physical system, and in future work, we will interface directly with the liquid substrate using the same approach.

We will also compare the optimized models we obtained here with those obtained when working with the physical substrate to evaluate how well our KdV model captures the behavior of the physical system across a range of parameterizations.
Although the KdV model is known to accurately capture the physics of shallow water surfaces, no mathematical model can perfectly describe a physical system.
This reality gap can be closed by introducing the physical system in place of the simulation in our evolution in materio framework, while using the findings from our expansive parameter search to narrow down the range of parameters we explore when working directly with the physical substrate.
The framework we applied here is not limited to use with hydrodynamic reservoirs but could also be applied to other physical systems whose dynamics have been well-described by physics-based models.

We demonstrated here that our evolutionary approach drastically improved the separation capabilities of the reservoir, compared to previous implementation with hand-selected parameters.
We also developed a generalized method of input encoding for continuous variables and demonstrated its effectiveness with a regression task.
Our experiments show that using our approach with the objective of maximizing separability for a set of training inputs produces models that can generalize to out-of-sample inputs in the same domain; however, our approach can be adjusted to also guide the reservoir toward the approximation property (i.e., producing similar outputs for similar inputs), which would improve out-of-sample generalization.
In future work, we aim to refine our approach to other similarly complex tasks, including how to select appropriate training data for a target task and confirming that the developed encoding method can be applied to other tasks.

\if@ACM@anonymous
\section{Conflict of Interest}
A. P., K. H., and S. N. declare no conflict of interest. S. S. and G. M. are employed by Apoha Ltd., a company dedicated to the development of biophotonic computational devices.
\fi

\begin{acks}
The authors are grateful to Giulia Marchiori Pietrosanti for valuable discussions on this project. This work was partially funded by the SOCRATES project (\grantsponsor{270961}{Research Council of Norway},, IKTPLUSS grant agreement \grantnum{270961}{270961}), the DeepCA project (\grantsponsor{286558}{Research Council of Norway},, Young Research Talent grant agreement \grantnum{286558}{286558}), and the Nordic Center for Sustainable and Trustworthy AI Research (NordSTAR), OsloMet internal initiative Ref.202237-100. The research presented in this paper has benefited from the Experimental Infrastructure for Exploration of Exascale Computing (eX3), (\grantsponsor{270053}{Research Council of Norway},, contract \grantnum{270053}{270053}).
\end{acks}


\bibliographystyle{unsrtnat}
\bibliography{sample-base}

\end{document}